\documentclass{article}

\usepackage{amsmath}
\usepackage{graphicx}
\usepackage{subfig}
\usepackage[figuresright]{rotating}
\usepackage{booktabs}
\usepackage{url}
\usepackage[square,numbers]{natbib}
\usepackage{todonotes}
\usepackage{authblk}

\title{FRANS: Automatic Feature Extraction for Time Series Forecasting}

\author[1]{Alexey Chernikov}
\author[1]{Chang Wei Tan}
\author[2]{Pablo Montero-Manso}
\author[1]{Christoph Bergmeir}

\affil[1]{Department of Data Science and Artificial Intelligence, Monash University, Melbourne, Australia}
\affil[2]{Disciple of Business Analytics, University of Sydney, Australia}

\date{}
\begin{document}

\maketitle

\begin{abstract}
	Feature extraction methods help in dimensionality reduction and capture relevant information. 
	In time series forecasting (TSF), features can be used as auxiliary information to achieve better accuracy. 
	Traditionally, features used in TSF are handcrafted, which requires domain knowledge and significant data-engineering work.
In this research, we first introduce a notion of static and dynamic features, which then enables us to develop our autonomous Feature Retrieving Autoregressive Network for Static features (FRANS) that does not require domain knowledge. The method is based on a CNN classifier that is trained to create for each series a collective and unique class representation either from parts of the series or, if class labels are available, from a set of series of the same class. It allows to discriminate series with similar behaviour but from different classes and makes the features extracted from the classifier to be maximally discriminatory.
	We explore the interpretability of our features, and evaluate the prediction capabilities of the method within the forecasting meta-learning environment FFORMA. Our results show that our features lead to improvement in accuracy in most situations. Once trained our approach creates features orders of magnitude faster than statistical methods. 
\end{abstract}

\section{Introduction}

Methods to extract characteristics from time series (also called time series features) have proven to be valuable tools in the forecasting space, to analyse time series and to ultimately produce better forecasts.
For example, in the prominent M4 competition, the second and third method relied heavily on feature extraction \cite{Makridakis:2020cm}.
In particular, the second-placed method, the hybrid statistical and machine learning model \textit{feature-based forecast model averaging (FFORMA)} \cite{MonteroManso:2019cd} is built with a meta-learner-based approach and uses statistical methods like Auto-Regressive Integrated Moving Average (ARIMA), Exponential Smoothing (ETS), Seasonal, Trend Decomposition using Loess (STL), and several others as base models. The method heavily utilises the \texttt{tsfeatures} \cite{Hyndman:1Va2jS34} suite of manually crafted statistical features. Here, the features extracted can be as simple as a mean or variance of a series, autocorrelations, or more complex features like the spectral entropy or degree and type of non-stationarity of a series. Those features are the major input used then by a meta-learner in FFORMA.
The meta-learner is an ensemble of gradient-boosted trees that learns to combine the forecast models, using the features as input, optimising a loss function designed for the model combination. 

While FFORMA was very successful in the M4 competition (\citet{Cawood2022} even show that adding ES-RNN \cite{Smyl:2018vu} to FFORMA's methods pool achieves state-of-the-art performance in the M4 competition dataset), the manual features are potentially inflexible in the sense that they may work well on the M4 dataset, but not necessarily on other datasets. As such, an automatic feature extraction step seems desirable. In the machine learning literature, there is a vast body of literature on automatic feature extraction, usually using autoencoders. In a time series forecasting context, for example, \citet{Laptev.2017} and \citet{10.1109/icdmw.2017.19} used a long short-term memory (LSTM) autoencoder as a feature extraction method. Later on these extracted features were used as exogenous variables to improve the forecasting accuracy of a neural network model.

However, autoencoders have certain drawbacks, such as that when using RNN autoencoders, the reconstructed series tend to be very smooth, and the autoencoders have difficulties to capture high-frequency information (see \cite{DONUT} for examples). Fundamentally, the problem here is in the reconstruction loss or the concept of the similarity between the original and the reconstruction. Most autoencoders use MAE or RMSE, but in time series, the usefulness of those measures often is limited, for example, when the relevant information is located in the high-frequency component of the series. Another example is a similarity up to phase shifts: two time series that are shifted in time by one step can look very different under MAE/RMSE, while in reality, they are nearly the same.

On the other hand, CNN-based autoencoders require all series to be of the same pre-defined length. One way to overcome this issue is to use a windowing approach, but then the features are only representative of a particular time series window, not for the whole time series, as is usually done with manual feature extraction. This issue is getting worse in cases when there are several time series per data source/class (e.g., IoT devices) -- autoencoders are naturally incapable of identifying the connection between the series of the same source and making appropriate adjustments to the feature creation mechanics among same-source series.

Thus, we propose a method for automatic feature extraction from time series in a forecasting context that is able to overcome these problems by using a data-driven notion of similarity. In particular, the main contributions of our paper are as follows: 

\begin{itemize}
	\item We present a notion of dynamic features that describe a time series at a certain moment in time, i.e., they are extracted from each series/window separately, and the notion of static features that describe a time series as a whole, across different series/windows. We analyse the differences of these two concepts and how machine learning methods can be used to extract those features.
	\item We present an autonomous self-supervised method for the extraction of static features from time series.
\end{itemize}

\noindent We are then able to show that the features extracted with our proposed method:
	\begin{itemize}
		\item contain more information in less volume than the current features extracted by statistical models
		\item surpass standard statistical handcrafted features and state-of-the-art methods of feature selection in time series forecasting in the FFORMA meta-learner environment
	\end{itemize}

	The remainder of the paper is organised as follows. Section~\ref{sec:relwork} discusses the related work, Section~\ref{sec:method} describes the methodology used, Section~\ref{sec:exper} provides the results of the experiments, and finally, Section~\ref{sec:concl} concludes the paper.

\section{Related work}
\label{sec:relwork}

Feature extraction from time series has a long tradition, and goes all the way back to the seminal work of \citet{Tukey.1982} where that author described a way to analyse multivariate data. The work brought several novel techniques to data examination such as the first use of a pictorial approach based on a graphical tool PRIM-9, and secondly the first use of extracting the hidden nonlinear structures to create lower dimension manifolds which could be depicted for the visual representation in PRIM-9 for further analysis. Though at the time not discussed as those, today those very structures could be called data features.
More recently, time series features have been used for time series forecasting and classification in various ways. 
Generally, depending on the origin, time series features can be divided into four types:
\begin{itemize}
	\item Handcrafted features based on domain knowledge and time series characteristics
	\item Semi-automatically chosen features from a previous set of handcrafted features
	\item Distance-based features
	\item Automatically extracted features
\end{itemize}

\noindent We discuss these four groups of features in the following sections.

\subsection{Handcrafted features and semi-automatically chosen features} \label{sec_hand_feats} 

In the forecasting space, handcrafted features have been used for decades.
\citet{Collopy.1992} were the first to use features as an indicator for the selection of the most appropriate forecasting algorithm by introducing a rule-based system for model selection.
However, the system was heavily dependent on manual analysis and input, hence \citet{Adya:2001uc} improved the system by eliminating judgmental feature coding and introducing automatic heuristics capable to detect a set of features to be used in RBF.

\citet{Hyndman:2015en} carefully crafted a set of features and used Principal Component Analysis (PCA) to detect anomalies in time series. Since the initial work, these features have become popular in their implementation in the \texttt{tsfeatures} package in R \cite{Hyndman:1Va2jS34}, and they have been used successfully for forecasting, e.g., in the FFORMA algorithm \cite{MonteroManso:2019cd}.

\citet{BenDFulcher:2014uo} gathered a long list of more than 9,000 handcrafted features from various scientific disciplines in one database and tested each one of them on time series classification (TSC) tasks. 
Those authors showed that it is possible to outperform the baseline of TSC on the UCR time series archive \cite{UCRArchive2018}, using a smaller feature set with relevant features.
Further interdisciplinary research \cite{Fulcher:2017vj} of numerous datasets from various fields made available time series analysis algorithms to turn time series into a set of features for classification, clustering, and other tasks.

While offering a host of handcrafted time series features, the work of \citet{BenDFulcher:2014uo} at the same time made it challenging to select the right features for specific tasks. In addition, using all the features makes it computationally expensive and oftentimes is not a feasible approach in real-world settings. \citet{Lubba:2019cw} consequently proposed the ``catch22'' features for TSC, which is a set of 22 features selected from the thousands of handcrafted features from the previous work of \citet{BenDFulcher:2014uo}.
In some cases, handcrafted features may significantly outperform other techniques \cite{Khan:it}. More often they are successfully combined with modern algorithms like deep neural networks \cite{Egede:2017gi}. There are also successful attempts in the literature to use handcrafted features for forecasting time series with LSTM networks \cite{Bandara2019Forecasting}.

Despite the ubiquity of handcrafted time series features, they have certain disadvantages:

\begin{enumerate}
\item They can be computationally costly to compute due to the high dimensionality, and they may not be discriminatory to certain domains.
\item Many of the features require the time series to have a certain minimal length.
\item The approach requires deep domain knowledge and significantly more work in feature engineering and therewith will be focused on a limited range of applications.
\end{enumerate}

\subsection{Distance-based features}

Distance-based methods of time series characterisation calculate the similarity between time series and use the resulting distances as an input for, e.g., a classification method. One of the most popular classification methods is the k-Nearest Neighbours algorithm, which despite its simplicity produces strong competitive results \cite{Ding:2008,10.1007/s10618-016-0483-9,10.1007/s10618-019-00663-x}.

Distance-based features can be local or global features. An example for local distance-based features are ``shapelets'' \cite{Ye:2009do}, which identify the parts of a time series representative of the class it belongs to. With shapelets, distances between shapelets and time series are calculated to obtain a distance  matrix. 
Next, dynamic time warping (DTW) is one of the most widely used similarity measures that warps a time series to better match the other time series. DTW has traditionally been used as a distance measure for a Nearest Neighbour algorithm (NN-DTW). \citet{Kate:2015hn} treats the DTW distance between time series as a feature that can be passed into a classification algorithm. Later on, DTW was used by \citet{Iwana:2020} in combination with a convolutional neural network (CNN) for time series classification. The method demonstrated state-of-the-art results on the UCI datasets however failed to achieve the same performance on the UCR datasets.

In terms of forecasting, DTW possesses important characteristics inherent to the method -- an invariance to elastic distortions leads to the invariance to the temporal localisations. That in turn makes forecasting with the method inaccurate unless certain modifications are applied, such as a differentiable timing error loss as proposed by \citet{Rivest.2020}. While solving the temporal invariance, that method was designed only for binary time series.
A method that is able to address the binary limitation and take into account both shape and temporal localisations is DILATE as presented by \citet{Guen:2019}. This machine learning method with temporally-constrained DTW uses a neural network with two losses: a shape loss and a temporal loss. The shape loss is based on DTW to calculate the structural shape dissimilarity between prediction and ground truth. The temporal loss is a modified Time Distortion Index (TDI) as presented by \citet{Frias.2017}.

\subsection{Automatic feature extraction}

The vast majority of automatic feature extraction approaches in the literature are based on autoencoders and use a sliding window method to cut the series into pieces. They then process those single pieces either in an unconnected way, or with a ``soft'' connection via the timesteps mechanism of an LSTM network.
In the following, we discuss the different methods used in this space.

\subsubsection{MLP and LSTM autoencoders}
Autoencoders have been used to extract features for many tasks including forecasting. \citet{Wang.2017} used a two-tiered architecture to forecast electricity prices.
The first tier (pre-training stage) consists of a denoising autoencoder which extracts features in the middle layer. In the second tier, the extracted features were used as an input in another deep neural network to provide forecasts. To accomodate time series of various lengths the approach is using a sliding window method. 
A so-called replacement autoencoder is another variation of a denoising autoencoder introduced by \cite{Anonymous:J4oVQt9T}, in combination with the sliding window approach those authors used their system to create in the middle layer features for distinguishing white-listed and black-listed data in privacy-preserving analysis for sensory data.

\citet{Gensler.2016} forecast the power output of solar powerplants using a similar 2-tier architecture. In the first stage, an autoencoder based on an MLP was used for feature calculations in the bottleneck layer. The second stage uses the encoding part to produce features which feed into the LSTM forecasting model.

A similar architecture was used by \citet{Laptev.2017} while attempting to solve the problem of forecasting of segments with high variance. 
It was also based on a 2-tier architecture, using LSTM in both autoencoder and forecaster. 
In the first stage, the LSTM autoencoder calculates time series features, which are used as input later for the second stage in the LSTM forecasting model. 
Features from additional data such as wind and temperature are used together with the data samples in the LSTM forecaster. 
Those authors then used a sliding window method to cut the data for the autoencoder in the way that each window covers appropriate intervals of each of the additional data sources. An autoencoder creates feature vectors for each data source, after that the final feature vector is aggregated (averaged) from the previously obtained feature vectors and finally inserted into the LSTM forecaster.

Finally, \citet{Bao:2017gd} implemented a 3-stage framework for financial time series: (1) the time series are firstly denoised with a wavelet transform, 
(2) the denoised time series are then fed into stacked autoencoders, and (3) finally, extracted features are fed into an LSTM for one-step-ahead prediction. 
The resulting model was more stable than both LSTM and an LSTM with data denoised with a wavelet transform (WLSTM).

\subsubsection{Convolutional autoencoders}

Initially created for images, CNNs have become popular in time series classification and forecasting because of their natural ability to create features by default as a part of the data workflow. The main difference in time series implementations is that modern implementations of convolution layers for time series are usually one-dimensional.
 
 One of the earliest papers in using a CNN in the time series domain tackles the problem of Human Activity Recognition \cite{o3n} and uses 2D CNNs, similar to typical image recognition networks such as ImageNet \cite{Krizhevsky:2017wl}, but with different convolution kernel sizes.
 A sliding window with a step of 3 is used to cut the time series into shorter lengths that can be used as input for the CNN model.
 Features produced by the network were then flattened and transferred into a fully connected layer.

\citet{Cirstea.2018} used 2 approaches, namely a convolutional recurrent neural network (CRNN) and an auto encoder CRNN (AECRNN), in a time series forecasting problem.
The former approach is a standard CNN without a flattening layer.
This made it possible to concatenate the network with the RNN layer after the merged pooling layer which contains all the features from all the univariate time series. 
AECRNN used an autoencoder approach, where the merged pooling layer containing the features was followed by deconvolution layers for creating the autoencoder.
At the same time the same merged pooling layer was also fed into the RNN to produce the forecasts. 
Therefore, this architecture creates the merged pooling layer with 2 outputs. Both architectures use a rolling window approach to split the input time series.

\subsubsection{Hybrid CNN-LSTM autoencoders}
 More complex 2D convolutional autoencoders have been used in \citet{Kieu.2018}. Their pipeline started with a 2-step enrichment process.
 First, statistical features were calculated from each sliding window extracted from a time series. These features were concatenated with the initial window data, thus producing enriched samples in the form of a matrix for each of the time series.
 Windows of a size $b$ overlapped each other by $b/2$. The result of this step is a stack of enriched time series.
 The second step works with the resulting data from the previous step: 2 enriched time series concatenated into one sample.
 Then the sample is used as input into a 2D CNN autoencoder which produced features. The result of such manipulations is a 3D matrix, where each time series contains $X$ 2D matrices of enriched samples. Logically, since a CNN autoencoder is able to capture dependencies only within a single matrix, no temporal dependencies within time series can be captured.
  Hence, an LSTM autoencoder is added. After that, the enriched samples and features from both 2D CNN and LSTM autoencoders are concatenated into one feature vector that is fed into a fully connected layer to produce the final output. Since a CNN autoencoder was used as the feature extractor, the extracted features were dynamic.

\section{Methodology}
\label{sec:method}

In the following, we first introduce the concept of static and dynamic features and then our methodology based on these concepts.

\subsection{Static and dynamic features}
\label{sec:dyn_stat_feats}

The handcrafted features discussed in Section~\ref{sec_hand_feats} are typically applied to a time series as a whole, and therewith accordingly describe the behaviour of the full time series, regardless of the current time frame. 
As such, the handcrafted features are normally used to create a (pseudo) unique image of a time series, a static representation. Some of the features of different time series might have similar values, however, the whole feature set distinguishes time series with different behaviour from each other. We call these features static features as their intent is to identify certain characteristics of a series as a whole, that do not change (much) over time. At the same time, it is important to note that not all of the handcrafted features are static -- as an example, autocorrelation might change significantly over time in certain time series, and thus in such cases, it cannot be called a static feature.

In this work we focus on the idea that one time series is one unique class, since in forecasting problems generally the series are not associated to classes. Therefore, to create a static representation of a series through static features we split the series via a sliding window method to create a library of snapshots of the series from different points in time. At the same time it is worth noting that our method of feature extraction is not limited by the approach ``one series-one class''. There are many cases where the concept to be captured is not a single time series. For example IoT devices that generate time series of a certain length in a certain period of time. In this case there will be no continuous time series but rather a set of separated series from the same source and of the same nature. In this case our method can be used to create the static representation of the whole set, since it is creating this representation out of pieces of information, which in this case will be every separate series in the set.

Thus, due to our focus on the concepts to be extracted being identical to single time series, without class labels present, for the research we use sliding window methods that are very popular when using machine learning on time series. This is mostly to address the potential changing lengths of time series, and to overcome limitations of the hardware and methods, where long time series may exceed the RAM capacity in complex deep neural networks.

Hypothetically, autoencoders trained with no sliding window, i.e., when the full time series fits into the model as well are capable to produce static features, however for the majority of cases fitting the whole time series into the model is not feasible due to hardware and other limitations as follows. For example, a simple LSTM autoencoder with input size of 384 data points and 3 encoding hidden LSTM layers and 3 decoding LSTM layers (384In-1536Units-768Units-384Units-FeatureLayer-384Units-768Units-1536Units-384Out) fails to fit in an 8Gb GPU. Therefore, in practice, many authors effectively window the time series by defining a maximum length of the series that is used. For example, \citet{DONUT} use a maximum length of 500 data points per time series.
Another major drawback of RNN/LSTM autoencoders in general is that with reasonable amounts of compression they tend to lead to poor reconstructions, in the sense that reconstructions are very smooth, and any high-frequent components of the signal are lost, limiting their ability to capture high-frequent behaviour of the time series in their features. Examples of this can be seen in \citet{DONUT}, (Figure 10, page 19, of that paper).

The use of sliding window techniques has important implications in many methods, in the way that no notion of time across the full time series in consideration is generated, beyond the windows.

\begin{figure}[htb]
\includegraphics[width=1.0\textwidth]{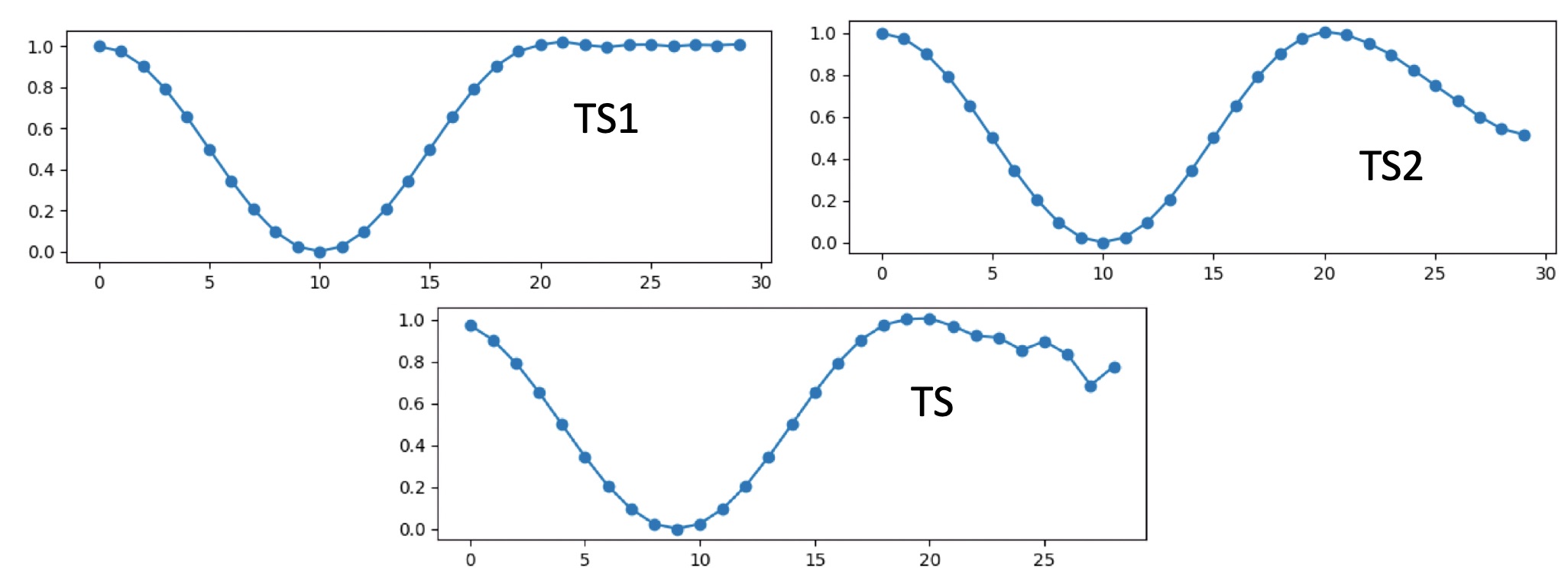}
\caption{An example of an autoencoder to approximate a sine wave. Top: Two time series with identical behaviour in the first 2/3rd of the series and different in the last 1/3rd of the series.
Bottom: Resulting restored series by a neural network autoencoder in both cases.}
  \label{fig:pic_sine}
\end{figure}

In particular, as autoencoders aim to output a reconstruction from their input, we observe that 1) in a windowed process, the autoencoder reconstructs the current window, and there is no connection to other windows from the same time series. And 2) the autoencoder creates an approximation among all similar windows regardless of which time series the windows originate from. As an autoencoder builds internal representations (aka features) in a middle layer (between encoder and decoder), these representations of similar windows of different time series are also similar and approximated. Thus, for two almost identical samples with, e.g., a difference in only one data point, the restored time series in that special data point will have the value which equals an average between values of this datapoint in the two windows. An illustration of this problem is given in Figure~\ref{fig:pic_sine}. We can see that the autoencoder generates the same features for both time series even though the time series are different.

Thus, any other time frames (windows) of a particular time series except the one being restored currently are not taken into account. At the same time, time series can be similar in certain time frames and vary significantly in others. This in turn makes an autoencoder with a sliding window capable of producing features which can describe only the current window and not the time series as a whole. This is one of the reasons why \citet{Kieu:9iD1fQQh} claim that ``autoencoder ensembles only exist for non-sequential data, and applying them to time series data directly gives poor results.''
In summary, with a sliding window technique the autoencoder is capable of providing averaged descriptions (features) of time series in a certain moment of time, and the averaging takes place across all time series with similar behaviour at some moment of time, which in turn dynamically changes through time. As such, the features obtained in this way are dynamic.

The two approaches of static and dynamic features present rather contrasting and oftentimes unrelated notions, however, in the literature they are usually not distinguished. To summarise, their main differences are as follows:

\begin{enumerate}
	\item Static features:
	\begin{itemize}
		\item Describe the behaviour of a time series as a whole.
		\item Are unique for each time series.
		\item Can be used for forecasting, classification.
	\end{itemize}
	\item Dynamic features:
	\begin{itemize}
		\item Describe the behaviour of a time series in a certain moment of time.
		\item Average similar time frames across time series.
		\item Compress the data with high accuracy.
		\item Can be used in Big Data for compression, denoising etc.
	\end{itemize}
\end{enumerate}

\subsection{Model intuition}
\begin{figure}[htb]
\includegraphics[width=1.0\textwidth]{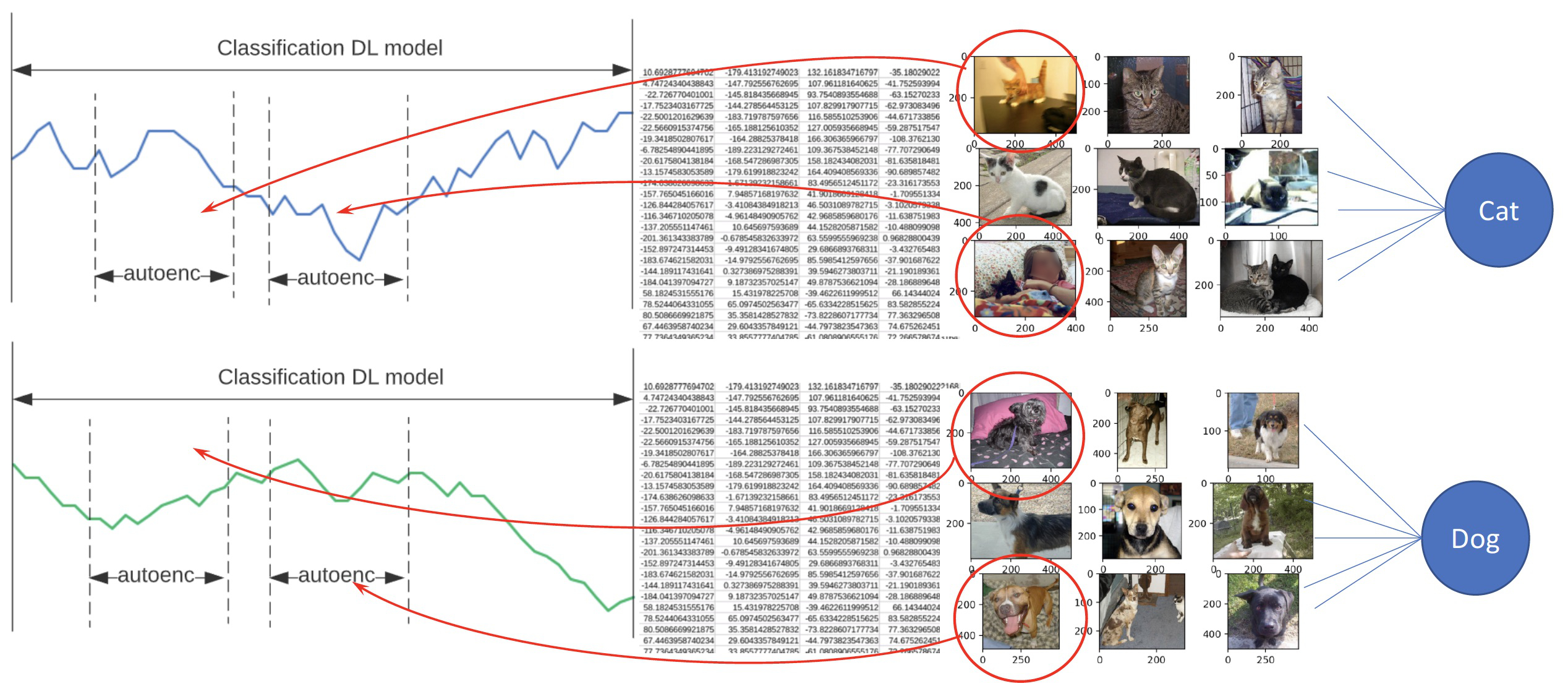}
  \caption{A time series representation for CNN feature extraction. On the right we see a standard CNN approach where each picture is a piece of information for a certain class. For example, each picture of a cat represents the class ``cat'', each picture of a dog represents the class ``dog''.}
  \label{fig:pic_intuition}
\end{figure}

The main aim of our work is to develop a neural network model architecture that allows us to extract static features from time series. For this, the neural network needs to develop a clear distinction between time series, and as such, it needs to take into account enough series/windows of each and every concept, which, depending on the addressed problem, can be a single time series or a class of time series.
This is in line with, for example, image classification tasks, where the feature space of an object is created from lots of samples of that object from many different images.
In our case, each series/window of a time series represents a separate ``image'' of the same time series (or group/class of time series). As such, a time series can be a ``class'', in the image classification analogy. See Figure~\ref{fig:pic_intuition} for an illustration.

Time series are sequential and will usually not have several other time series representing the same class, therefore we define a sliding window as one unit of information for the class that represents the particular time series. This makes it possible to combine the representation of the class from the fragments of the series, which makes it more robust to handle new, unknown input. At the same time unlike the autoencoder, where windows are not connected to a certain class and therefore features for the slightly different windows of different time series are averaged and similar to each other, our proposed approach allows us to make use of such little differences by forming unique features for windows having very similar data but different classes. This is possible as we use a supervised method and specific indications of the class. In particular, instead of autoencoders, we use simplified 1D-CNN-based models, designed for time series \cite{IsmailFawaz:2019ek}.

\subsection{Model architecture}
For the feature extraction, we use the model presented by \citet{IsmailFawaz:2019ek} for time series classification tasks as a base. We make certain modifications to make it suitable for feature extraction. 
Firstly, we insert a feature layer that is used for the feature extraction before the output. The dimensions of the layer equal the number of features we want to extract.
The model is trained as a multi-class classifier on time series windows as input. The labels represent the time series (e.g., all windows from the first time series in the dataset have ``TS1'' as their label, and so on).

Secondly, instead of the Categorical Cross-entropy loss function used in the original paper we are using the Sparse Categorical Cross-entropy. The change is caused by the fact that a standard one-hot encoded label of the Categorical Cross-entropy represented by a vector with length of the number of the classes, i.e., class 7 in 10 multiclass labels will look like [0,0,0,0,0,0,1,0,0,0], which is acceptable for the problems with classes up to few hundreds. At the same time, e.g., the largest dataset we consider in our experiments is the M4\_Monthly dataset that contains around over 40,000 time series (and therewith classes), which determines the label of the according length vector. Such labelling does not fit into the RAM of the GPU and accordingly we change it to the Sparse Categorical Cross-entropy.

\begin{figure}[htb]
	\centering
	\subfloat[]{
		\includegraphics[width=0.74\textwidth]{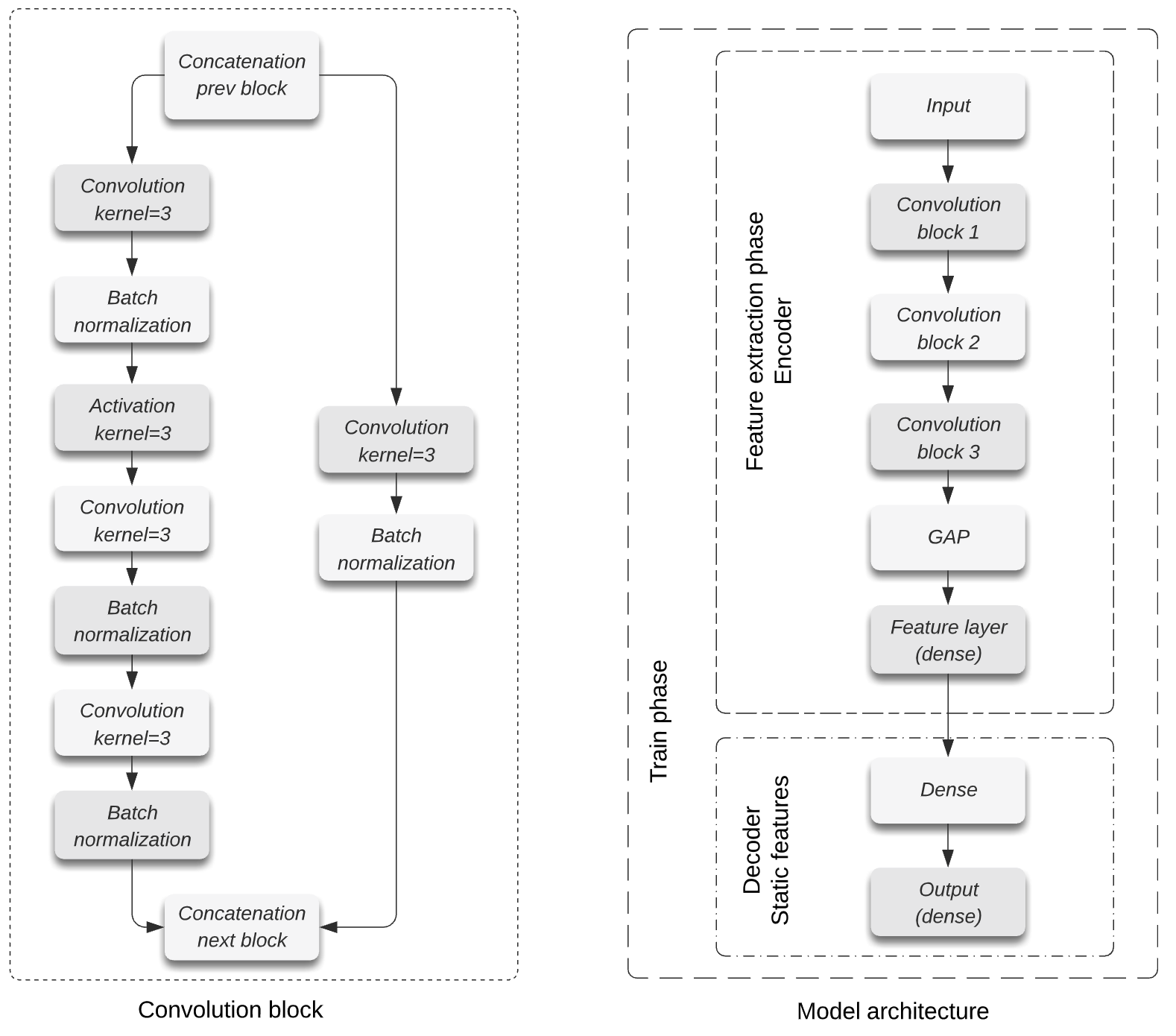}}
	\subfloat[]{
		\includegraphics[width=0.24\textwidth]{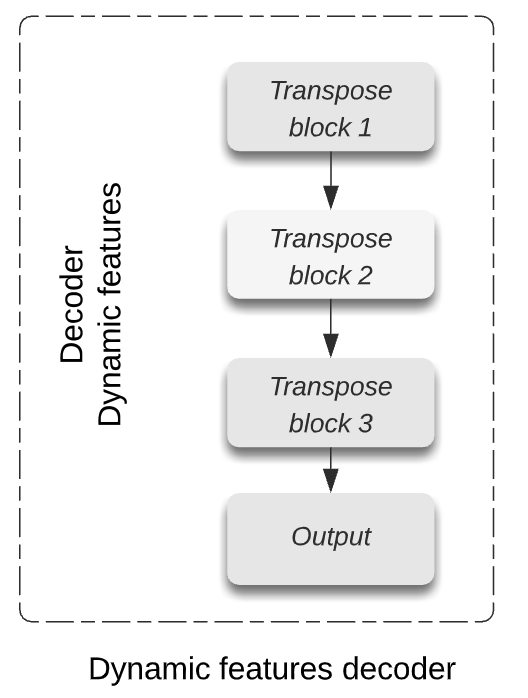}}
\caption{Feature extraction neural network. (a) The encoder and the decoder for the static feature extraction. (b) The decoder for the dynamic feature extraction.}
\label{fig:mod_mine}
\end{figure}

After training the model, we remove the top dense layer (output layer) from the model, and use the second-to-last layer directly as a layer that outputs features. The network can now be used to predict features for every window across all time series in the dataset.
The final architecture is presented in Figure~\ref{fig:mod_mine}.

Furthermore, as the main aim of the architecture is to extract static features that do not change across the time series, we further stabilise the features by calculating the mean and medoid of the features across all windows, which is used afterwards as a static feature vector for the time series in consideration (see Section~\ref{sec:eval_feat_qual} for an analysis).

As such, the average representation of the class (series) is conducted in our method in the same way it has been done in modern image classification tasks with CNN, started by \citet{Krizhevsky:2017wl}

\subsection{Incorporating Automatic Features into FFORMA}

FFORMA~\cite{MonteroManso:2019cd} is a meta-learning forecasting method based on statistical models from the \texttt{forecast} package \cite{Hyndman:2015fr} in R, such as Na\"{i}ve, Auto.ARIMA, Theta, ETS, and others, as base models for calculating predictions. The method then uses XGBoost \cite{Chen:uy} as a meta-learner trained on features extracted from the time series, to predict weights for each base model that are then used for a weighted averaging to produce the final forecasts. 

The approach proved to be 10\% more accurate than the model selection method, stable to the changes in the base model errors (less than 1\% drop in accuracy, when one base method is removed) and not dependent on the loss function of the prediction \cite{MonteroManso:2019cd}.

To extract features from the time series which can be fed to the meta-learner for consequent weight generation, the method utilises static time series features from the package \texttt{tsfeatures} \cite{Hyndman:1Va2jS34}. For every time series 42 static features are calculated and the resulting 2D matrix is used in the meta-learner as an input for the training stage.

We modify the FFORMA approach, where instead of the \texttt{tsfeatures} package for the feature extraction we use our deep feature set consisting of 16 features. We keep the rest of the methodology identical to highlight the impact of the features. The initial advantages of our method are a 2.5 times reduction of the feature space and the capability of processing short series, which is unfeasable for some of the features from \texttt{tsfeatures}. Furthermore, FFORMA and arguable also \texttt{tsfeatures} were created considering series from the M3 and M4 competition. Our automatic feature generation method is more flexible and adjust to other datasets with other characteristics. Thus, the method is able to yield higher forecasting accuracy, as demonstrated in Section~\ref{sec:FFORMAres}.

\section{Experiments}
\label{sec:exper}

In this section, we first describe the datasets used for evaluation (Section~\ref{sec:dataset}).
Then, in Section~\ref{sec:FFORMAres} we show that our automatically generated features are able to improve forecasting accuracy, by using them within the FFORMA framework on a total of 5 datasets.
And finally, in Section~\ref{sec:eval_feat_qual} we delve deeper into the characteristics of the extracted features by presenting an analysis and justification of their value and meaningfulness.

\subsection{Datasets}
\label{sec:dataset}

As the main aim of our paper is to present automatic features that are useful in a forecasting scenario, namely within the FFORMA algorithm, we focus on datasets commonly used in the forecasting literature. In particular, the dataset of the influential M4 forecasting competition~\cite{Makridakis:2020cm} contains 100,000 series with different frequencies, different lengths, and from different application domains. As the original FFORMA method was developed for this competition and won second place, it is a natural choice of dataset. Apart from that, as our aim is to show how our method is more flexible and adapts better to other datasets, we use other additional datasets, as follows. Table~\ref{tab:datadesc} presents summary statistics of all datasets used.

\begin{description}
\item[CIF2016\_Monthly]
 The Computational Intelligence in Forecasting 2016 (CIF2016) competition dataset contains 72 monthly time series originating from simulations and from the banking domain \cite{Stepnicka:ue}.
 \item[NN5] The NN5 competition dataset contains 111 daily time series with data covering daily cash withdrawals from ATMs in the UK \cite{nn5}.
 \item[Ausgrid\_Weekly] The Ausgrid ``solar home electricity dataset'' by the Australian electric operator managing the distribution grid in Sydney and nearby areas. We use a weekly version of the dataset from \citet{godahewa2021monash}.
 \item[Traffic\_Weekly] Traffic data from the Caltrans Performance Measurement System (PeMS) across all major metropolitan freeways of the State of California. We use a weekly version of the dataset from \citet{godahewa2021monash}.
\end{description}

\begin{table}[htb]
\centering
\begin{tabular}{lrrr}
\toprule
Set name &   Num of Series &  Min. Length &   Max. Length \\
\hline
\midrule
M4\_Yearly &  23000 &   19 &   841 \\
M4\_Quarterly &  24000 &   24 &   874 \\
M4\_Monthly &  48000 &   60 &  2812 \\
M4\_Daily &   4227 &  107 &  9933 \\
M4\_Hourly &    414 &  748 &  1008 \\
CIF2016\_Monthly &     72 &  120 &   120 \\
Ausgrid\_Weekly &    299 &  156 &   156 \\
Traffic\_Weekly &    862 &  104 &   104 \\
NN5 &    111 &  791 &   791 \\
\bottomrule
\end{tabular}
\caption{Summary of the characteristics of the used datasets.}
\label{tab:datadesc}
\end{table}

\subsection{Data partitioning and model evaluation}

The M4 dataset comes with pre-defined training and test partitions that we use. All other datasets were partitioned by choosing a block of data from the end of the series as the test set, the rest of the series as training set. We choose the test sets to have a size of 24, 7, and 8 data points for hourly, weekly, and monthly series, respectively.

As error measure, we use the symmetric mean absolute percentage error (sMAPE) as presented by \citet{smape_mark}, for a definition see Equation~\ref{equ:smape}. Here, $t$ is the current time step, $n$ is the size of the test set, $y_t$ is the actual value from the test set, and $\hat{y}_t$ is the corresponding forecast.

\begin{equation}
\textit{sMAPE}=\frac{200}{n} \sum_{t=1}^{n}\left(\frac{|y_t-\hat{y}_t|}{|y_t|+|\hat{y}_t|}\right)
\label{equ:smape}
\end{equation}

The major advantage of the sMAPE is that it is scale invariant, which makes it possible to use it for comparison of forecast performances between datasets of different nature and scales. On the other hand sMAPE has certain limitations. When the ground truth and the predicted value equals 0, sMAPE will have its maximal value of 200. Another specific of the sMAPE is its asymmetry -- it penalises underestimation more than overestimation (see, e.g., \citet{hansika}).

Nonetheless, the sMAPE is commonly used in time series forecasting tasks, and in particular it was used in the M4 competition \cite{Makridakis:2020cm}, and therefore enables us to straightforwardly compare our approach to the original FFORMA on that dataset.

\subsection{Forecasting accuracy results}
\label{sec:FFORMAres}

Table~\ref{tab:res} shows the results of forecasting accuracy across the comparison methods. We compare our proposed approach with the original FFORMA method (FFORMA\_orig in the table), and all base models from the original FFORMA method (for details of the base models refer to \citet{MonteroManso:2019cd}). We note that the base models contain standard forecasting methods such as ETS, THETA, TBATS, an automatic version of ARIMA, and also a feed-forward neural network (NNETAR), so that we consider the used benchmarks a representative set of state-of-the-art univariate forecasting methods.

We can see from Table~\ref{tab:res} that in 8 datasets out of 10 our approach demonstrates better performance than the original FFORMA. In the remaining 2 datasets, the margin by which FFORMA wins is small. 
Furthermore, we are able to show that our method outperforms all the base models on all datasets.

Finally, another advantage of our method is that it is not limited by the length of the series. In the yearly subset of the M4 dataset, the full \texttt{tsfeatures} set cannot be extracted in approximately 4,000 series out of 23,000. As such, in more than 25\% of the series the full feature set cannot be extracted on this dataset, as it is dominated by short series.

\begin{sidewaystable}
\centering
\begin{tabular}{lrrrrrrrrrr}
\toprule
 &  M4\_Daily &  M4\_Hourly &  M4\_Monthly &  M4\_Quarterly &  M4\_Weekly &  M4\_Yearly &    NN5 &  Traffic &  Ausgrid &  CIF2016 \\
 \hline
\midrule
Naive2          &     3.065 &     18.383 &          15.256 &        11.013 &      8.103 &     15.737 &      -- &           -- &              -- &               -- \\
RandomWalkDrift &     3.171 &     43.465 &          15.477 &        11.426 &      9.484 &     13.474 &      -- &           -- &              -- &               -- \\
SeasonalNaive   &     3.742 &     13.912 &          15.988 &        12.519 &     14.517 &     15.737 &  26.715 &       15.594 &          31.158 &           10.493 \\
Auto.arima           &     3.203 &     14.080 &          13.434 &        10.418 &      7.004 &     14.595 &  25.010 &       13.164 &          26.634 &            7.409 \\
ETS             &     3.127 &     17.307 &          13.525 &        10.290 &      8.727 &     14.648 &  22.087 &       12.817 &          29.930 &            6.596 \\
NNETAR          &        -- &     13.735 &          15.431 &            -- &         -- &         -- &  24.013 &       15.046 &          26.824 &            8.933 \\
STLM            &        -- &     16.612 &          17.126 &            -- &         -- &         -- &  23.000 &           -- &          26.448 &               -- \\
TBATS           &     3.002 &     12.442 &          12.941 &        10.187 &      7.302 &     14.100 &  21.285 &       13.269 &          24.918 &            6.959 \\
THETAF          &     3.071 &     18.138 &          13.012 &        10.313 &      7.833 &     13.905 &  22.594 &       12.834 &          30.150 &            8.319 \\
FFORMA\_orig      &     3.060 &     11.670 &          \textbf{12.741} &         9.848 &      6.838 &     13.011 &  \textbf{21.083} &       12.350 &          24.601 &            6.688 \\
Proposed method       &     \textbf{3.020} &     \textbf{11.622} &          12.806 &         \textbf{9.846} &      \textbf{6.729} &     \textbf{12.950} &  21.084 &       \textbf{12.208} &          \textbf{24.410} &            \textbf{6.556} \\
\bottomrule
\end{tabular}
\caption{sMAPE comparison between baseline methods and new Deep Learning feature extraction method}
\label{tab:res}
\end{sidewaystable}

\subsection{Analysis of feature quality}
\label{sec:eval_feat_qual}

In this section, we focus on the largest and most diverse dataset in our experiments, namely the M4\_Monthly dataset, which has 48,000 time series. We show analyses around the stability and quality of our features.

To be stable (and therewith representative for the whole time series) the features obtained from the different windows of one time series should have similar values, i.e., be located close to each other in feature space. At the same time, features from one time series should be placed farther from another time series in feature space, and the closer the distance between clusters formed by windows of certain time series, the more similar these time series should be. The opposite should also be correct: the farther the distance between clusters the less similar the time series are.

\begin{figure}[htb]
\includegraphics[width=0.8\textwidth]{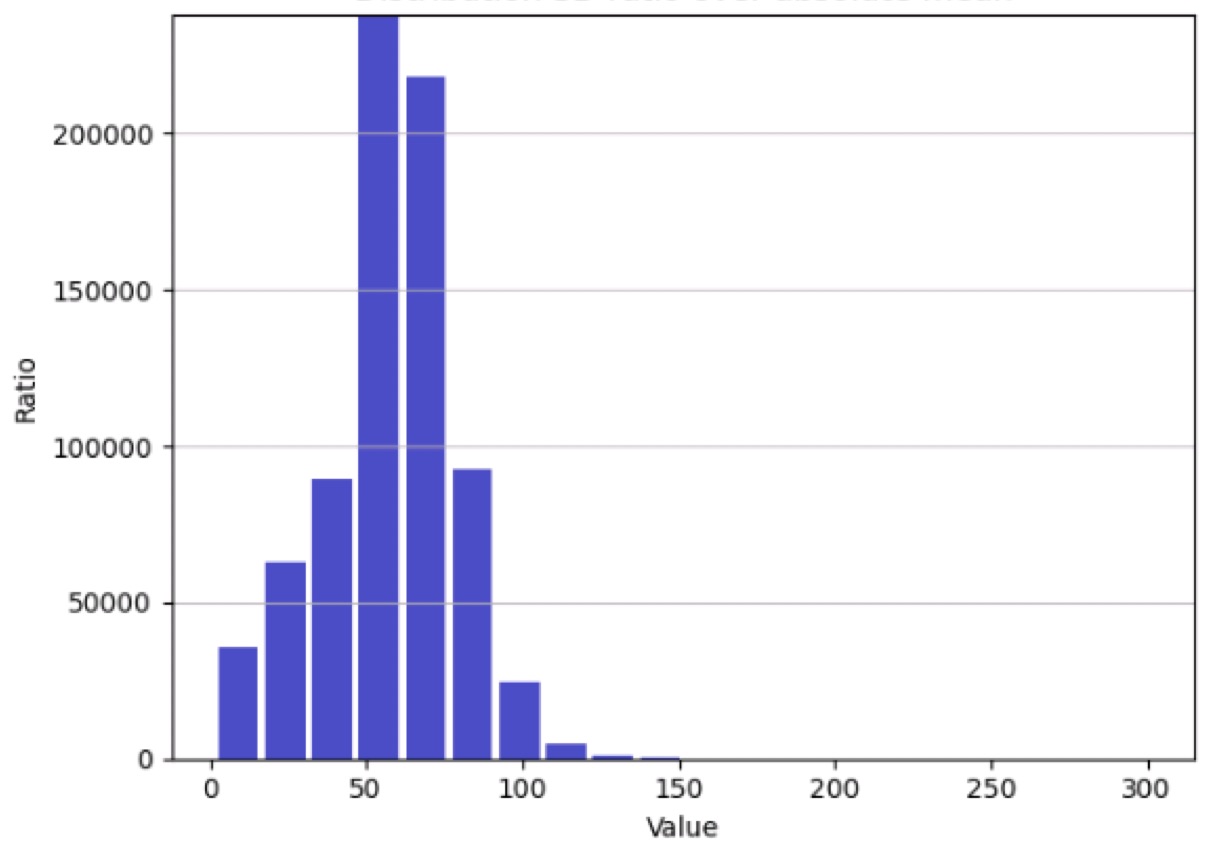}
\caption{Distribution of the variability of the features. The features generated by the network for each series are normalized by their absolute mean, and the standard deviation of all the features is calculated. The plot shows the histogram of the standard deviation divided by the mean of features for all series in the M4 dataset.}
  \label{fig:feat_sd}
\end{figure}

The histogram in Figure~\ref{fig:feat_sd} plots the 
ratio of standard deviation over absolute mean across all features. We can see that the standard deviation of a feature across all windows of the same time series can be several times larger than its mean value. As such, the features appear not as stable as we would desire. 
A solution to this problem is to calculate the mean across features to create static constant features.
Then, we need to ensure that the averaged features are closely related to the features of the windows. In particular, if the feature represents the time series it should be located in a similar place in the feature space as other windows of the time series. 

Finally, nonuniformity of the whole feature space would signal that features can be clustered and describe different types of time series behaviour. To evaluate our method along the above-mentioned dimensions, we use dimensionality reduction and clustering techniques, as discussed in the following sections.

\begin{figure}[!tbp]
  \centering
  \begin{minipage}[b]{0.475\textwidth}
    \includegraphics[width=\textwidth]{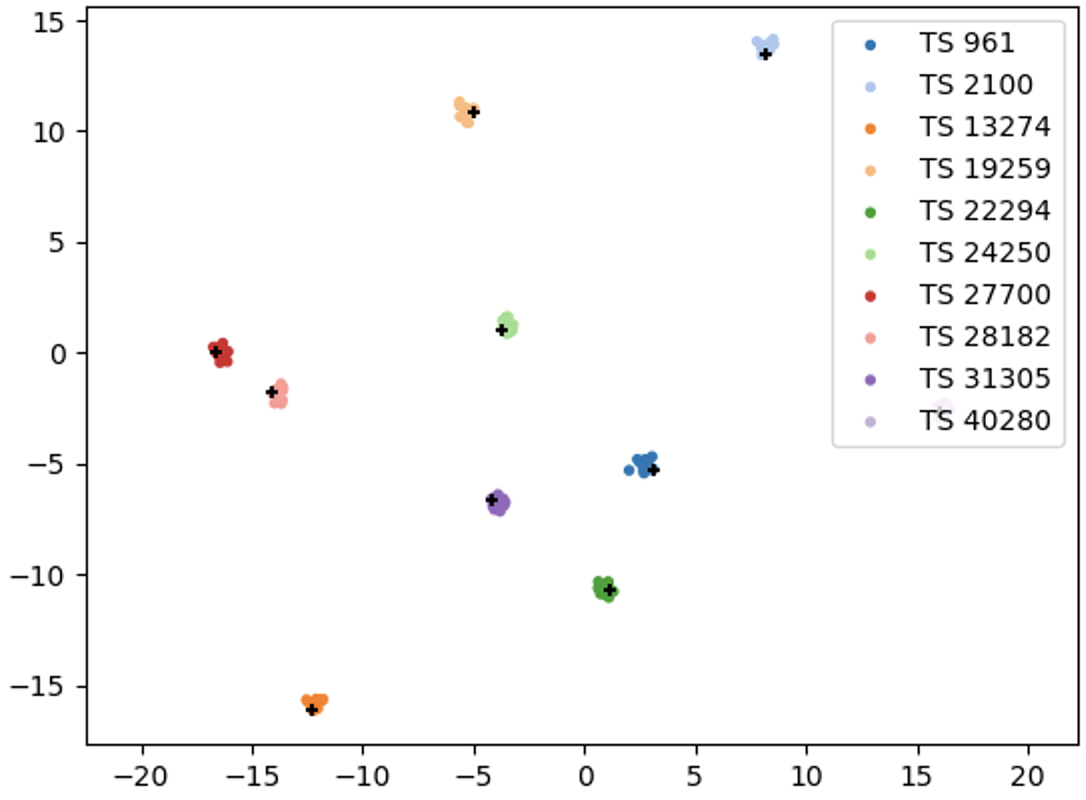}
  \end{minipage}
  \begin{minipage}[b]{0.475\textwidth}
    \includegraphics[width=\textwidth]{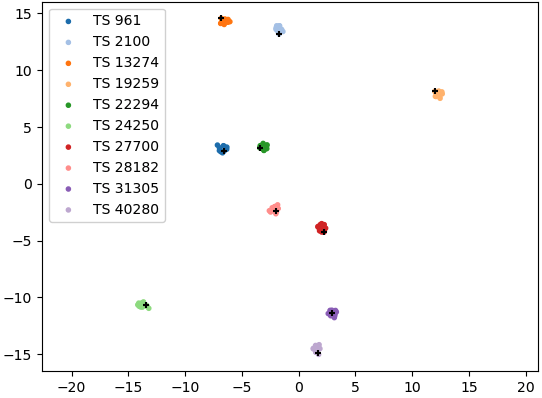}
  \end{minipage}
  \caption{UMAP representation of extracted features, 10 random time series with 10 windows each, M4\_Monthly dataset in colour for each series, black cross are the averaged features/medoids per each time series. Left -- centroid approach, right -- medoid approach.}
  \label{fig:feat_avg}
\end{figure}

\begin{figure}[htb]
\includegraphics[width=1.0\textwidth]{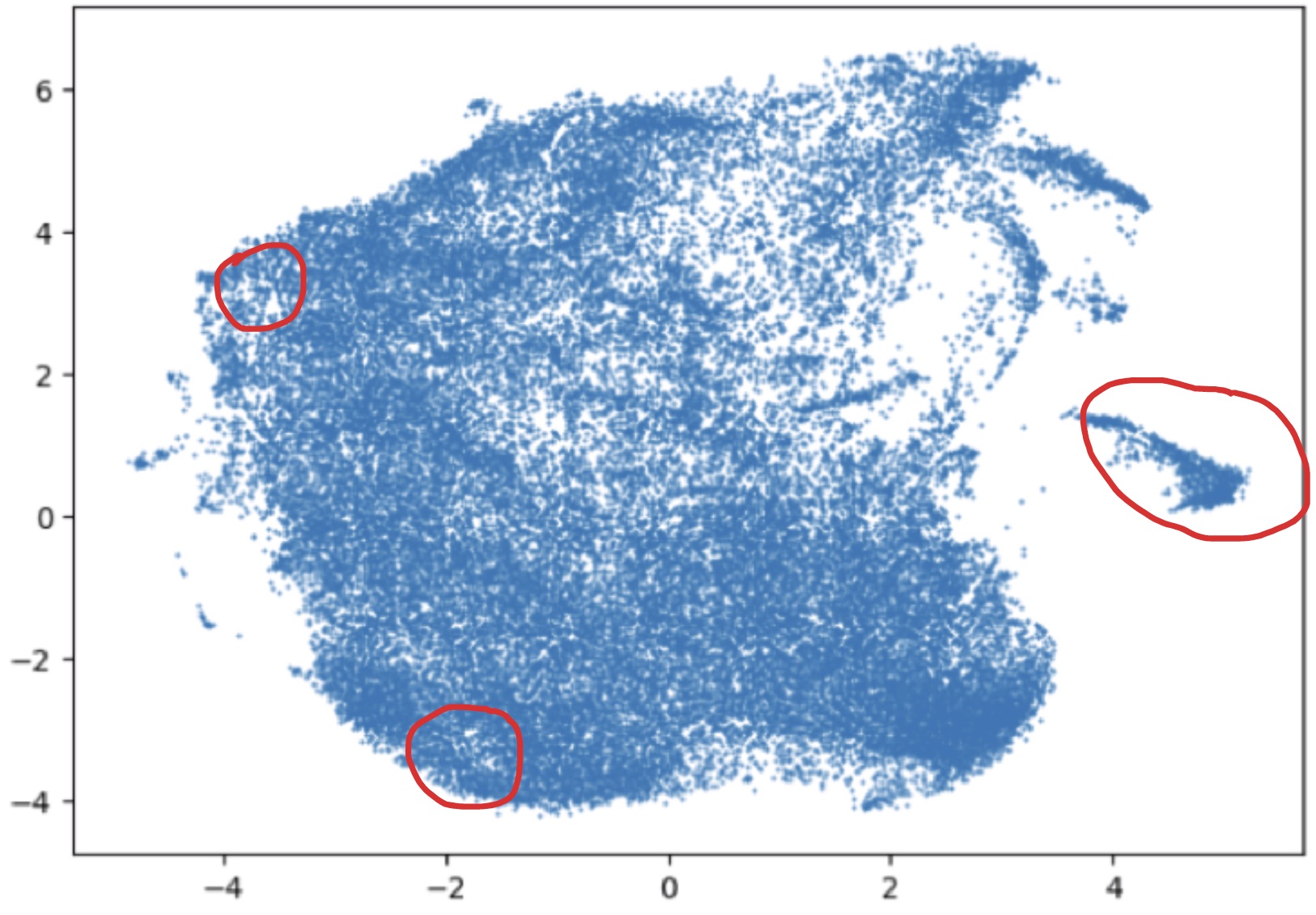}
  \caption{M4\_Monthly dataset, UMAP 2 component analysis of averaged features, 2D projection. The space contains the averaged features of the whole dataset. The main idea is that the behaviour of a time series is reflected by its averaged features, therefore features with similar behaviour should be located close to each other in the space. To check this, we chose time series from areas approximately equidistant from each other (encircled with the red shapes) and analyse the behaviour of time series in each of the selected areas further (see Figure~\ref{fig:clusters}).}
  \label{fig:dimRed2D}
\end{figure}

\begin{figure}[htb]
	\centering
	\subfloat[Region around the centre (x:0.5, y:4.9)]{
		\includegraphics[width=\textwidth]{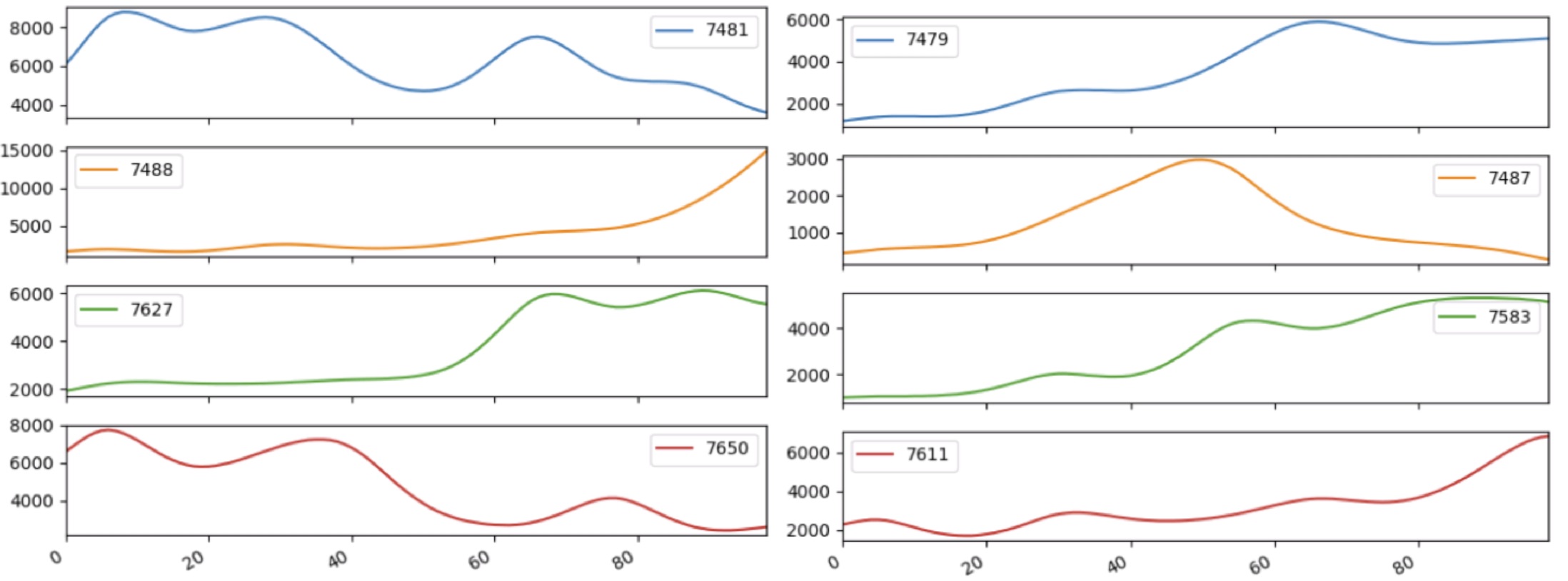}
		\label{fig:cl0549}}
	\qquad
	\subfloat[Region around the centre (x:3.5, y:-3.9)]{
		\includegraphics[width=\textwidth]{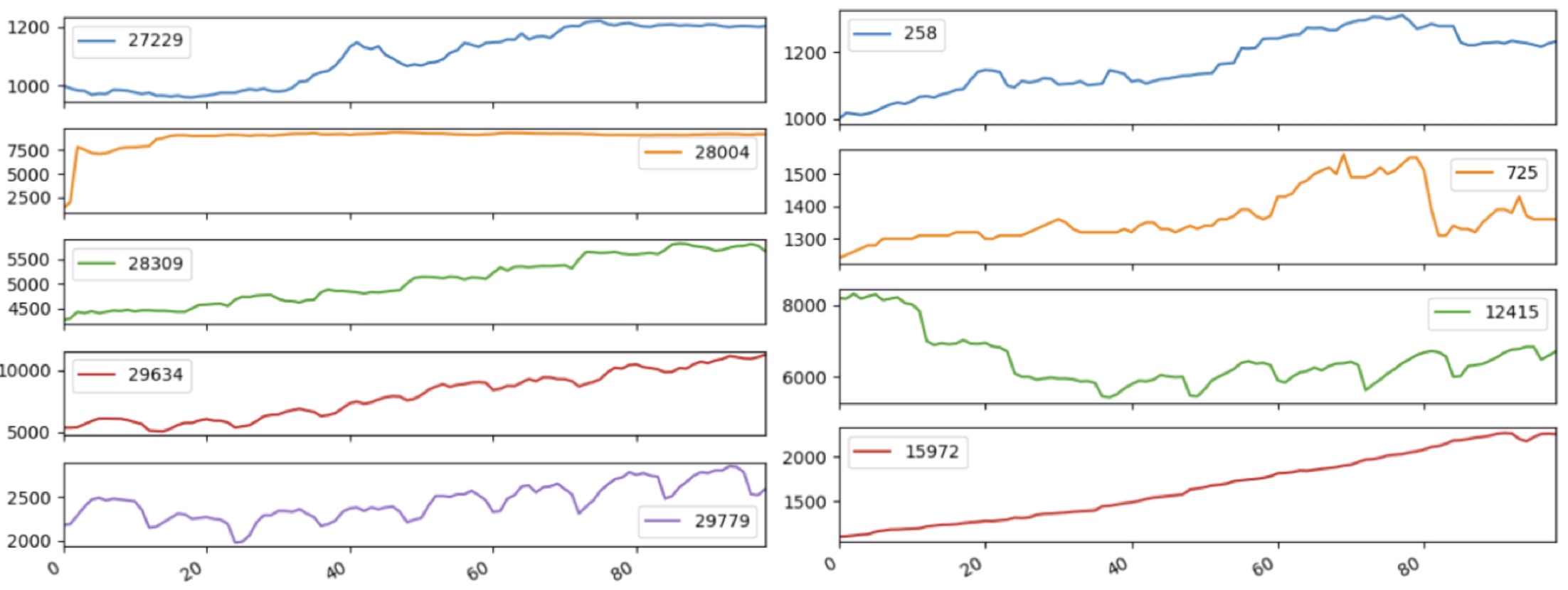}
		\label{fig:cl3539}}
	\qquad
	\subfloat[Region around the centre (x:-3.0, y:-2.0)]{
		\includegraphics[width=\textwidth]{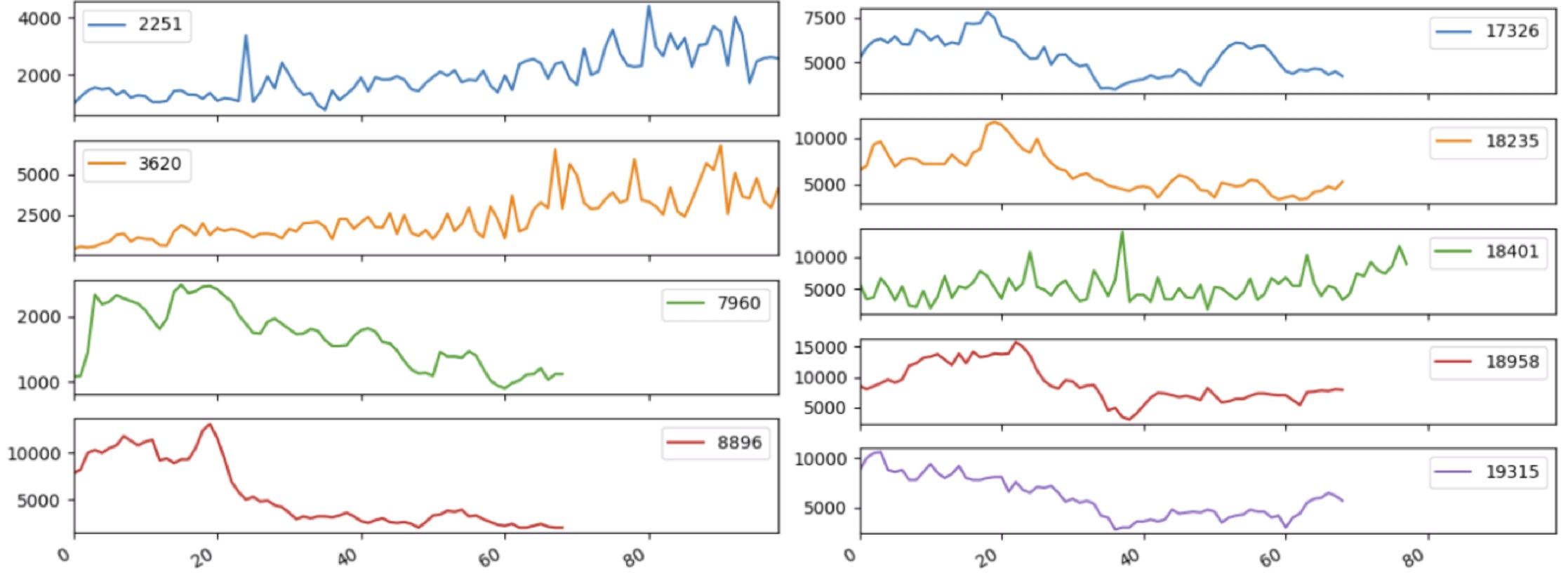}
	\label{fig:cl3020}}	
	\caption{Example time series from different regions of the feature space.
		TOP CLUSTER: Low frequency/smooth series.
		MID CLUSTER: periodic and trended
		BOTTOM CLUSTER: High frequency series}
	\label{fig:clusters}
\end{figure}

\begin{figure}[htb]
	\centering
	\subfloat[The most similar time series TS3448 and TS46685]{
		\includegraphics[width=\textwidth]{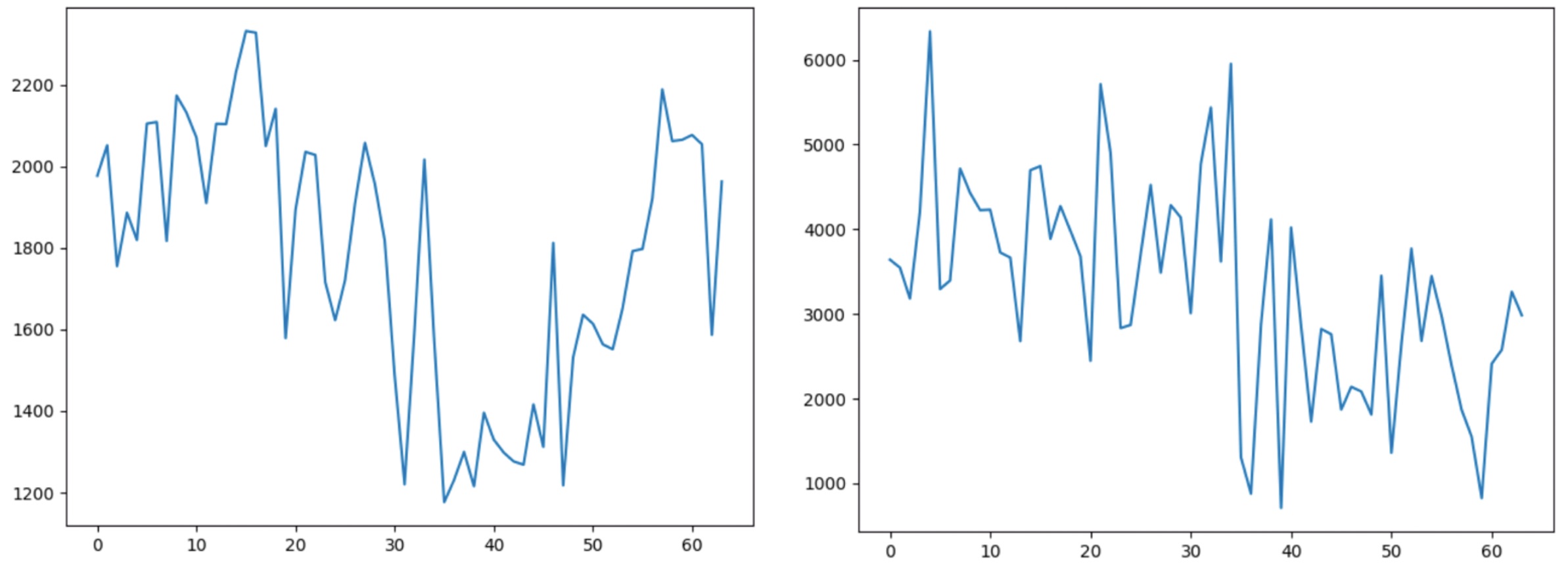}
		}
	\qquad
	\subfloat[The least similar time series TS7582 and TS37703]{
		\includegraphics[width=\textwidth]{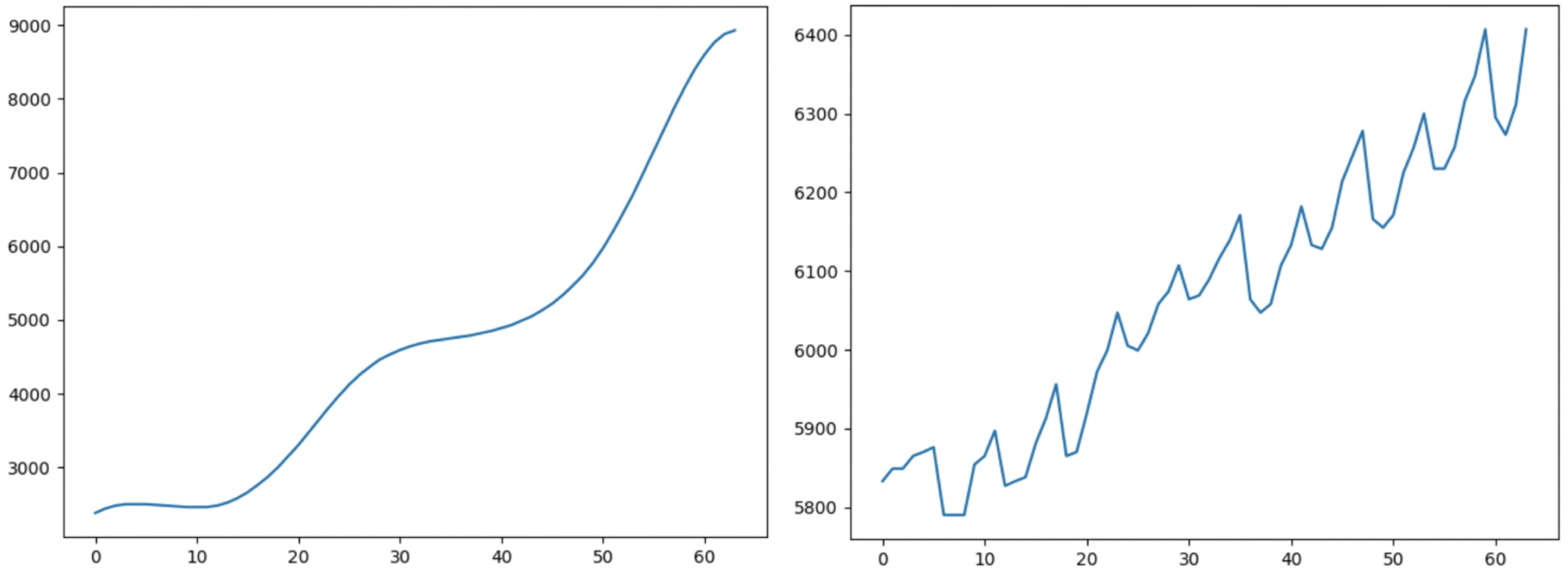}}
		\caption{The most and the least similar series from the M4\_Monthly dataset, according to the UMAP 2-components dimensionality reduction analysis and feature averaging.}
		\label{fig:dimRedDissimTS}
\end{figure}

\subsubsection{Dimensionality reduction analysis} 

We use dimensionality reduction techniques to illustrate nonuniformity and the existence of certain clusters in the feature space, as well as how the mean of the features per series behaves to the features of the windows.
In particular, our process is as follows:

\begin{itemize}
	\item We randomly choose 10 windows per time series and obtain features from these windows of all time series (480,000 windows in total)
	\item We process the feature sets of all chosen windows and the mean of features by UMAP and t-SNE dimensionality reduction methods
\end{itemize}
	
Based on this general process, we want to confirm at first that the features are relatively close to each other and that the mean of features is a good representation of a time series (to show static correctness).
As the dataset is large, we use random subsampling to make the results easier to understand and more illustrative.
Figure~\ref{fig:feat_avg} shows an example where we have randomly chosen 10 time series for visualisation. The features have been projected into 2D space using the UMAP dimensionality reduction technique.
We can see that in all observed cases all the features of each time series are located in close proximity of each other, including the mean. 

For the analysis of nonuniformity, Figure~\ref{fig:dimRed2D} shows a representation of the features in 2D space, using again the UMAP dimensionality reduction technique. We can see that the feature space is not homogeneous, and has some structure. To confirm that the different areas in the plot describe the behaviour of time series of certain types, we chose 3 areas distant from each other. The centres chosen in particular are (0.5, 4.9), (3.5, -3.9), and (-3.0, -2.0).
Figure~\ref{fig:clusters} illustrates that all 3 inspected regions have behaviours different from each other, yet similar behaviours inside a region.

This analysis indicates that the extracted features present good static features that describe the time series behaviour and could be used for clustering in the high-dimensional space to preserve as much information as possible. As an example, in Figure~\ref{fig:dimRedDissimTS} we can see the most and the least similar series according to the analysis.

\subsubsection{Clustering}

In this section we perform clustering of the time series features we extracted from the M4\_Monthly dataset to further investigate the capability of the extracted features in describing time series behaviours. We performed clustering using different methods, namely K-means++, Gaussian mixture, partitioning around medoids, and DBSCAN. All methods result typically in clusterings with 20-35 clusters. As the results across the methods are similar, we report in the following the experiments with K-means++ clustering.

\begin{figure}[htb]
\centering
\includegraphics[width=0.6\textwidth]{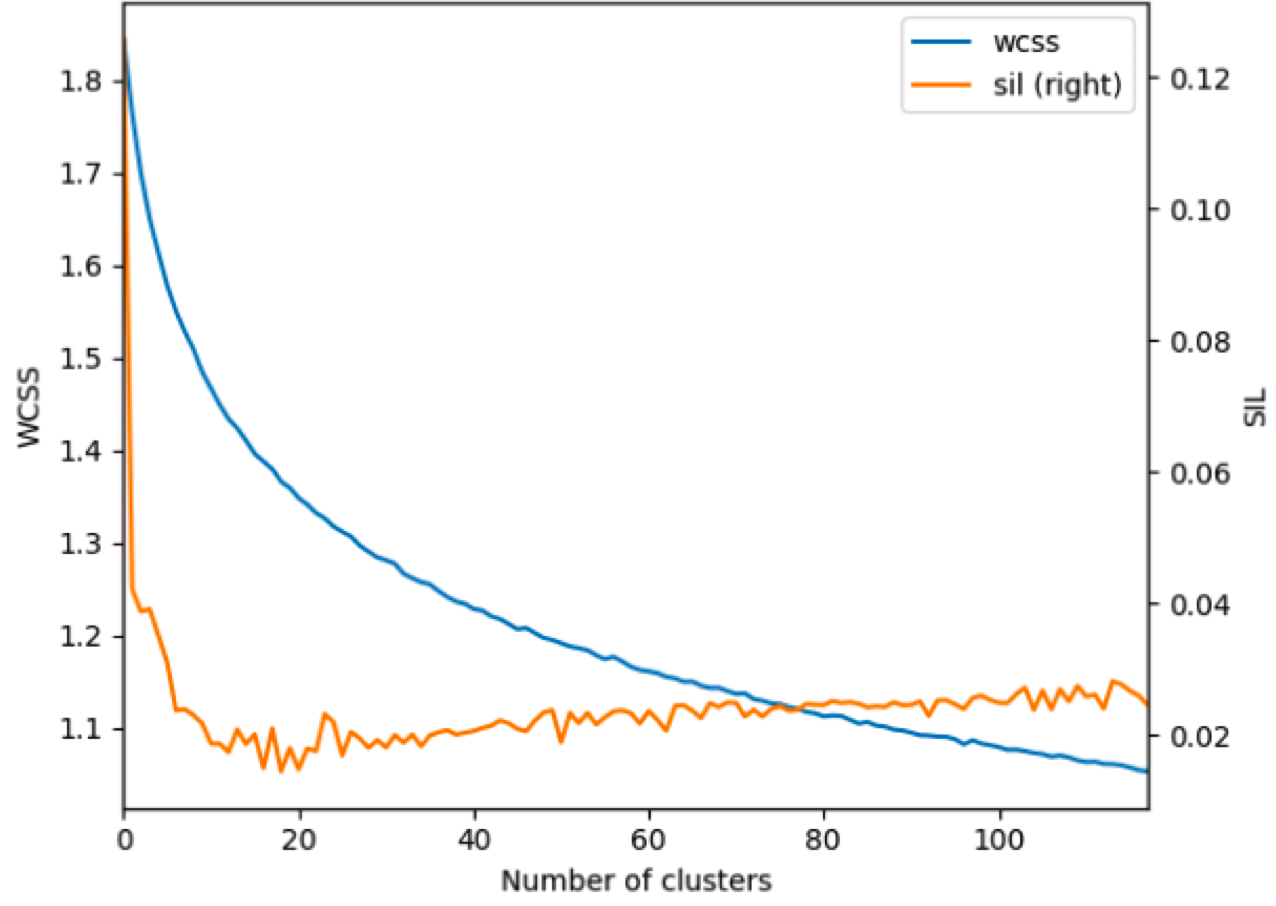}
  \caption{K-means++. Elbow method to select the amount of clusters.}
  \label{fig:cl_elb}
\end{figure}

The elbow method shows that the optimal number of clusters performed by K-means++ clustering on Euclidean distance lays in the range from 19 to 35 (Figure~\ref{fig:cl_elb}).
 The silhouette score provides small peaks in the 19 and 23 clusters region, the latter is slightly higher. 
 
\begin{figure}[htb]
\centering
	\subfloat[Mean of the clusters]{
\includegraphics[width=\textwidth]{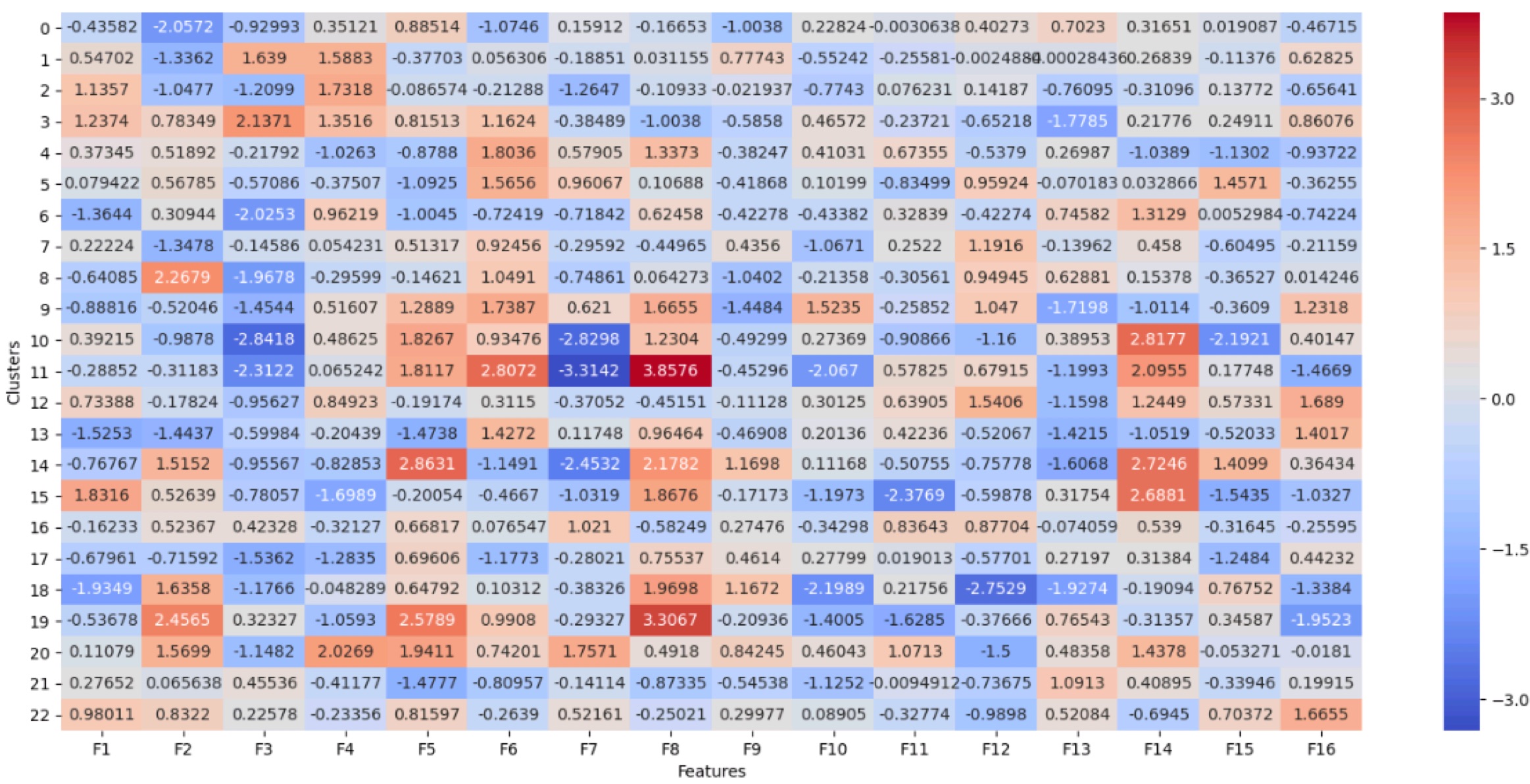}
}
	\qquad
		\subfloat[Standard deviation of the clusters]{
\includegraphics[width=\textwidth]{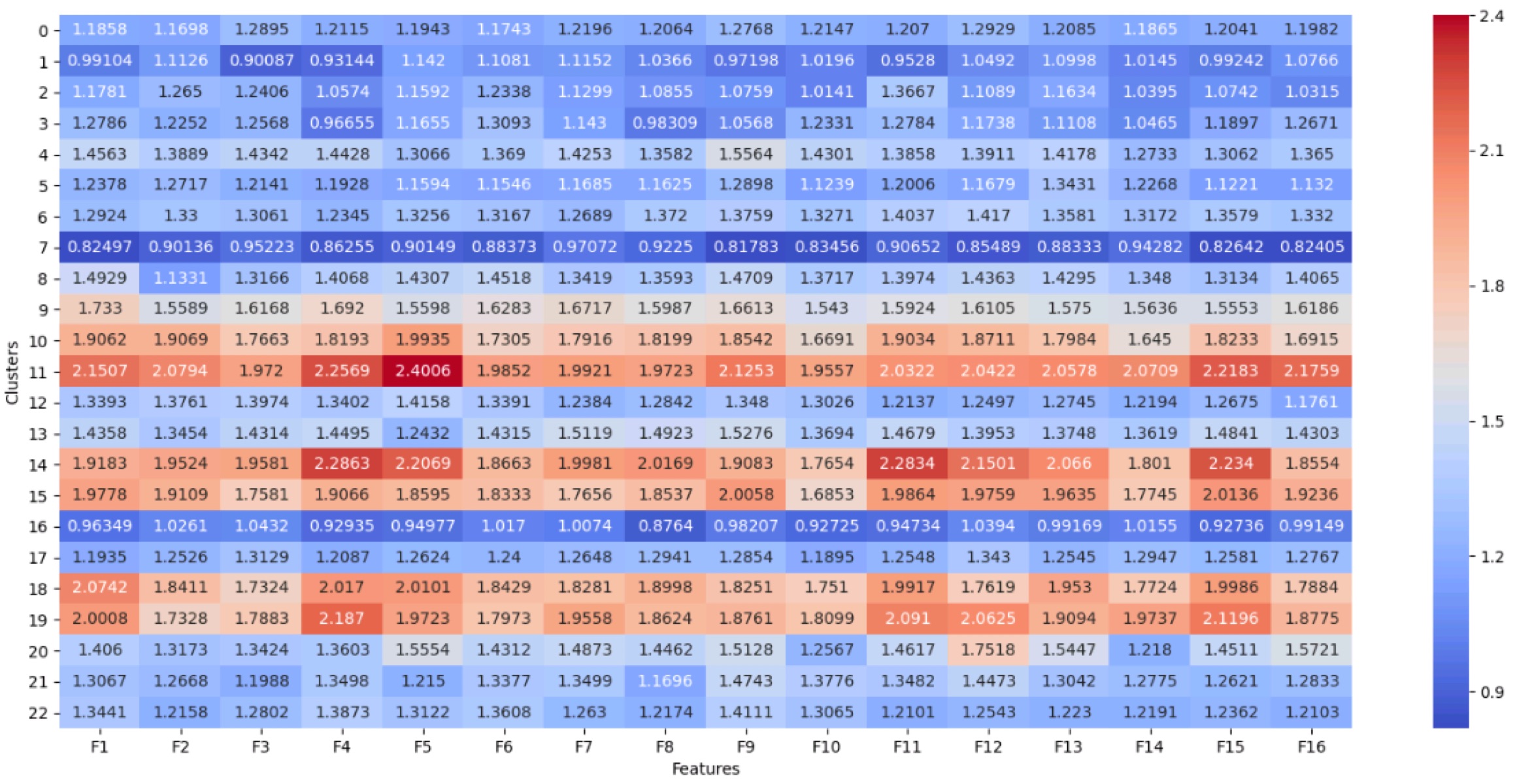}		
		}
  \caption{K-means++. Feature means and standard deviations per cluster on the M4\_Monthly dataset.}
  \label{fig:cl_km_means}
\end{figure}

\begin{figure}[htb]
	\centering
	\subfloat[Cluster 1, low SD]{
		\includegraphics[width=0.6\textwidth]{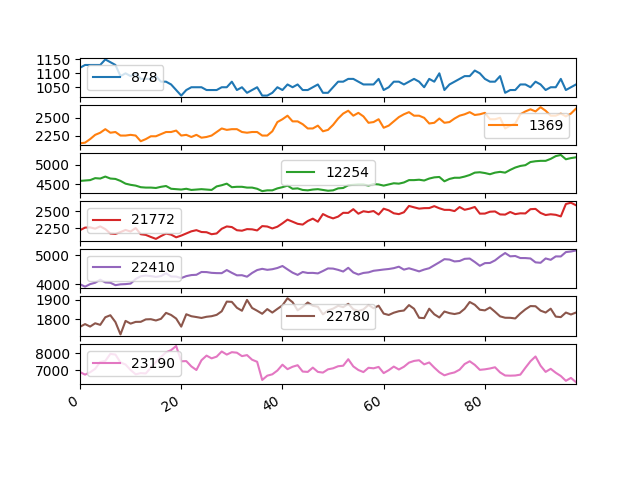}}
	\subfloat[Cluster 11, high SD]{
		\includegraphics[width=0.6\textwidth]{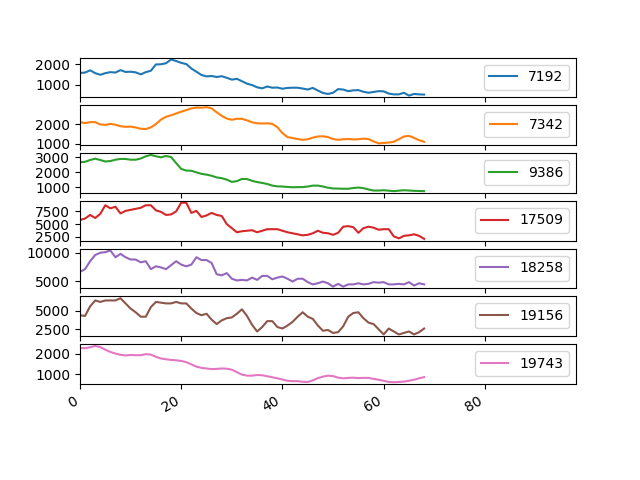}}
	\qquad
	\subfloat[Clusters 7, low SD]{
		\includegraphics[width=0.6\textwidth]{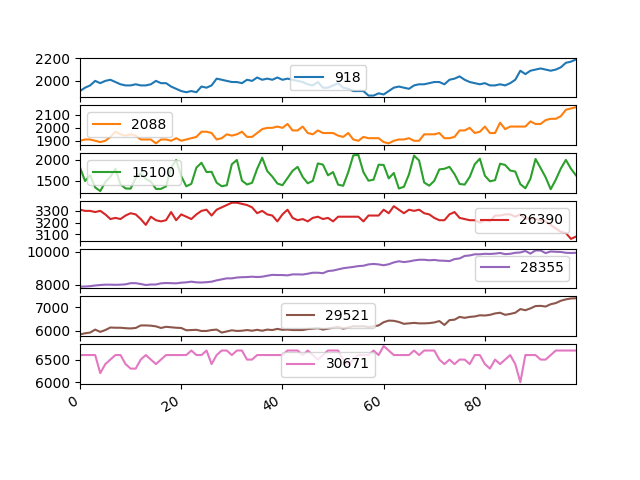}}
	\subfloat[Clusters 14, high SD]{
	\includegraphics[width=0.6\textwidth]{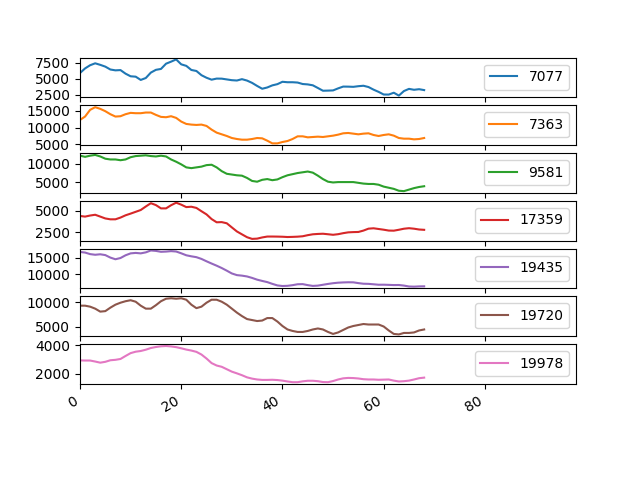}}
	\caption{Plots of randomly chosen series from clusters with lowest and highest SD.}
\label{fig:clust_sd_plots}
\end{figure}

 Dividing the M4\_Monthly dataset into 23 clusters using the extracted features gives us the table of cluster means and standard deviations (SD) shown in Figure~\ref{fig:cl_km_means}. We can observe that some features like F3, in general, have smaller mean values than others. In terms of the SD, we can see how the features depend on the cluster, namely Cluster 1 and 7 have the lowest SD in all the features, as opposed to Clusters 11 and 14 which have the highest SD. In Figure~\ref{fig:clust_sd_plots} we can observe time series from these clusters.

\subsection{Performance}
Apart from the feature sensibility, another major characteristic of any feature extraction method is the computational cost of the algorithm. In our experiments, we found that the \texttt{tsfeatures} method extracts features in 7252.9 seconds for the M4\_Monthly dataset (the largest dataset in our experiments). Our method, once trained, takes 10.1 seconds for the same task on the same hardware, a standard single-processor Intel i5 workstation, which is almost 720.25 times faster. The training takes 3727.0 seconds, which is nearly 50\% less time compared to \texttt{tsfeatures}. At the same time, for the production environment, it is important to notice that the training is done just once and after that, the feature creation takes seconds for as many repeats as needed, whereas with \texttt{tsfeatures} the second and the following feature creation processes will take the same amount of time as the initial one.

\section{Conclusions}
\label{sec:concl}

In this paper we have shown that time series features, in general, can be split into two different groups -- dynamic and static features, where the dynamic features describe the behavioural momentum of the series and the static features describe the general behaviour of the time series across all the data points. Since in forecasting tasks the most valuable features are the ones that describe the general behaviour of the series, the static features are the most appropriate choice in a forecasting context. 
Thus, we have developed a static feature creation method based on a CNN network, where we define a surrogate classification task where each time series is a separate class. After that, using the extracted features, we have been able to show that they can be used within the FFORMA forecasting method to achieve higher accuracy than with the manually defined features currently used in the algorithm. At the same time, our static features do not require any domain knowledge and are extracted in a fully automated fashion. Finally, in an explanatory analysis of the features using dimensionality reduction and clustering, we have illustrated their interpretability and reliability.

As future work we consider applying the static features in classification tasks. Regarding time series forecasting, the combination of static and dynamic features holds the promise to be even more beneficial than the static features alone. 

\section*{Acknowledgements}

This research was supported by the Australian Research Council under grant DE190100045 and Monash University Graduate Research funding.

\bibliographystyle{plainnat}
\bibliography{TSFeaturesP1}

\end{document}